\documentclass{article}
\usepackage{ijcai22}

\usepackage{times}
\usepackage{soul}
\usepackage{url}
\usepackage[hidelinks]{hyperref}
\usepackage[utf8]{inputenc}
\usepackage[small]{caption}
\usepackage{graphicx}
\usepackage{amsmath}
\usepackage{amsthm}
\usepackage{booktabs}
\usepackage{algorithm}
\usepackage{algorithmic}
\usepackage{amsfonts}
\usepackage{todonotes}
\usepackage{dsfont}
\usepackage{xcolor}
\usepackage{mathtools}
\usepackage{wrapfig}
\usepackage{soul}
\usepackage{cancel}
\newcommand\Ccancel[2][black]{\renewcommand\CancelColor{\color{#1}}\cancel{#2}}
\usepackage{array}
\usepackage{multirow}
\usepackage{float}

\newcommand{\FT}[1]{\textcolor{black}{#1}}
\newcommand{\delete}[1]{}
\newcommand{\deleteag}[1]{}
\newcommand{\AG}[1]{\textcolor{black}{#1}}

\title{Logically Consistent Adversarial Attacks for Soft Theorem Provers}

\author{
Alexander Gaskell$^{1,2}$\and
Yishu Miao$^{1,2}$\and
Francesca Toni$^1$\And
Lucia Specia$^1$
\affiliations
$^1$Imperial College London 
\\
$^2$ByteDance\\
\emails
\{alexander.gaskell19, f.toni, l.specia\}@imperial.ac.uk, yishu.miao@gmail.com
}

\begin{document}

\maketitle

\begin{abstract}
    Recent efforts within the AI community have yielded impressive results towards ``soft theorem proving'' over natural language sentences using language models.
    We propose a novel, generative adversarial framework for probing and improving these models' reasoning capabilities. Adversarial attacks in this domain suffer from the \textit{logical inconsistency} problem, whereby perturbations to the input may alter the label. Our \textbf{L}ogically consistent \textbf{A}d\textbf{V}ersarial \textbf{A}ttacker, \textbf{LAVA}, addresses this by combining a structured generative process with a symbolic solver, guaranteeing logical consistency. Our framework successfully generates adversarial attacks and identifies global weaknesses common across multiple target models. Our analyses reveal naive heuristics and vulnerabilities in these models' reasoning capabilities, exposing an incomplete grasp of logical deduction under logic programs. Finally, in addition to effective probing of these models, we show that training on the generated samples improves the target model's performance.
\end{abstract}

\section{Introduction}
    \label{sec:intro}
    
    Recent research highlights that language models are surprisingly adept at memorising and recalling factual information when trained on internet-scale corpora \cite{lm_as_kb}. Less clear, though, is their ability to deduce new conclusions by manipulating and combining this stored knowledge. The NLP community explore this via the publication of natural language benchmarks requiring logical reasoning \cite{ReClor}. While neural models are increasingly improving on these tasks, their reasoning capabilities can prove brittle and shallow upon closer examination \cite{helwe2021reasoning}.
    
    \begin{figure}
        \centering
        \includegraphics[width=0.48\textwidth]{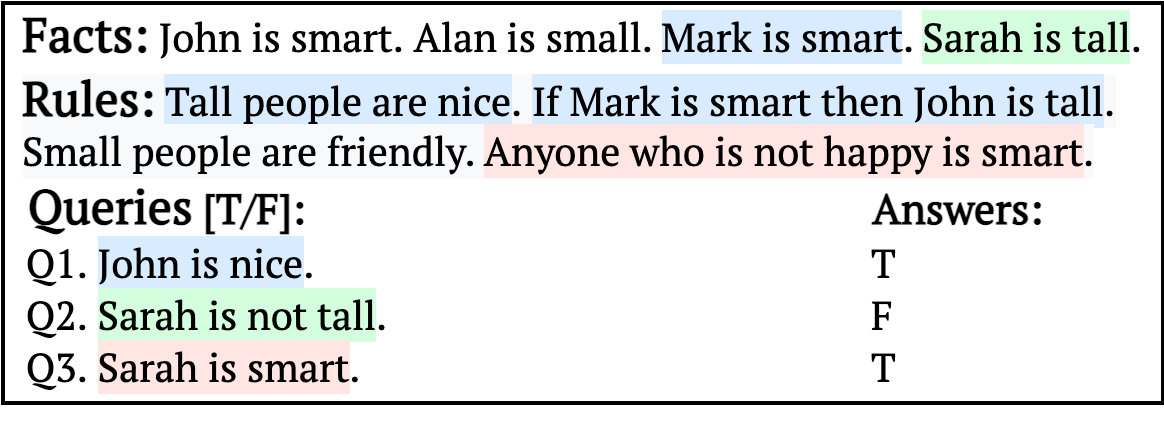}
        \vspace{-0.6cm}
        \caption{A (simplified) illustration of the \textit{RuleTakers} dataset 
        \protect\cite{ruletaker}. 
        The task is to predict whether a query is entailed by a set of facts and rules.
        }
        \label{fig:sample}
    \end{figure}
    
    Existing studies seemingly demonstrate that language models can act as ``soft theorem-provers'' (STPs) over natural language sentences.
    \cite{ruletaker} demonstrate that Transformers \cite{attention_is_all_you_need} can learn to predict whether a natural language query is logically entailed by a collection of natural language facts and rules, as illustrated in Fig.~\ref{fig:sample}. Their results appear impressive, showing near-perfect performance on the original task, demonstrating robustness to syntactic and lexical variations and generalization beyond the training set. Despite this, an important question remains: have these systems learned the desired reasoning processes, or are they dependent on shallow heuristics and shortcuts?
    
    Adversarial machine learning \cite{adversarial_ML} offers a natural inroad into this question. It is widely used within NLP to probe model robustness and defend against adversarial attacks \cite{hotflip,bert-attack}. A successful attack minimally perturbs an input such that: 1) its semantics 
    is preserved, and 2) the \delete{target} model alters its prediction. 
    The first challenge is usually addressed by assuming that ``small'' perturbations will not meaningfully impact semantics. 
    However, previous works focus on lexical semantics while we are interested in logical semantics.
    The
    assumption that ``small'' perturbations will not impact logical semantics  is unduly strong 
    as logical entailment is sensitive to any perturbations, causing standard methods to generate \textit{inconsistent} attacks by inadvertently flipping the label.
    \delete{This we refer to as the \emph{logical inconsistency} problem. }
    Considering Fig.~\ref{fig:sample}, existing methods may perturb \textit{``Mark is smart''} to \textit{``Mark is intelligent''}, but
    this attack is inconsistent for Q1 as it is no longer entailed after the perturbation. \FT{To solve this \emph{logical inconsistency} problem, our contributions are as follows:}\delete{we propose \textit{logically consistent attacks}, whereby the perturbed sample's label faithfully reflects its new entailment
    .}
    %
    %
    \delete{This paper proposes a novel framework for probing and improving STPs using logically consistent attacks. Our main contributions are as follows:}
    \begin{enumerate}
        \item We
        \FT{propose \textit{logically consistent attacks}, whereby the perturbed sample's label faithfully reflects its new entailment, and} \delete{introduce the notion of logically consistent adversarial attacks and demonstrate its importance by showing }
        \FT{show} that standard methods  produce \FT{logically} inconsistent attacks.
        \item We propose LAVA (\textbf{L}ogically consistent \textbf{A}d\textbf{V}ersarial \textbf{A}ttacker): a black-box, generative adversarial framework to select, apply and verify adversarial attacks on STPs. We demonstrate that the vanilla version significantly outperforms standard methods, and can be further improved via a simple best-of-$k$ decoding enhancement.
        \item LAVA exposes naive heuristics and global weaknesses common across multiple STPs, such as a flawed usage of quantified rules. Training on the adversarial samples improves the target model's performance.
    \end{enumerate}
    Our implementation is available at \url{https://github.com/alexgaskell10/LAVA}.
    
\section{Background: Soft Theorem-Proving}
    \label{sec:NLD}
    
    \cite{ruletaker} examine neural systems' capacities to conduct automated reasoning using language and introduce the \textit{RuleTakers} dataset to this end. They demonstrate that Transformers can act as ``soft theorem provers'' over natural language by solving a logical deduction task. Their objective is to determine whether a natural language claim, the \textit{query}, is ``entailed'' by a set of rules and facts, the \textit{context}. 
    
    To generate the dataset, the authors first synthesise stratified 
    logic programs, amounting to sets of rules and facts.
    \deleteag{In the simplest case, f}Facts are atomic sentences and rules are implications that have, as a body, conjunctions of atoms and, as the head, atoms. The authors also consider more general versions of rules and facts, where atoms may be negated.
    All rules  can be propositional 
    or (implicitly) universally quantified.  
    Then, rules and facts \deleteag{(with and without negation) } are converted to natural language representations using templates (see Fig.~\ref{fig:sample} for an example\deleteag{ of the result of this conversion}). 

    To generate labels for queries in the dataset, the authors use the closed world assumption and an interpretation of negation as  negation-as-failure (NAF), adapting the standard semantics for stratified logic programs to allow a symbolic solver to label whether a query is entailed by a context.
    We use $\models$ to represent this entailment.
    We illustrate NAF in Fig.~\ref{fig:sample}, Q3.,
    as Sarah cannot be shown to be happy so she is smart.
    
    The dataset is subdivided on a $[0..5]$ range according to the number of deductive steps required (\emph{proof depth}), approximating sample difficulty. Proof depth is closely related to proof \textit{length}, the number of context sentences in the proof.

\section{Logically Consistent Adversarial Attacks}
    \label{sec:model}
\begin{figure}
        \centering
        \includegraphics[width=0.48\textwidth]{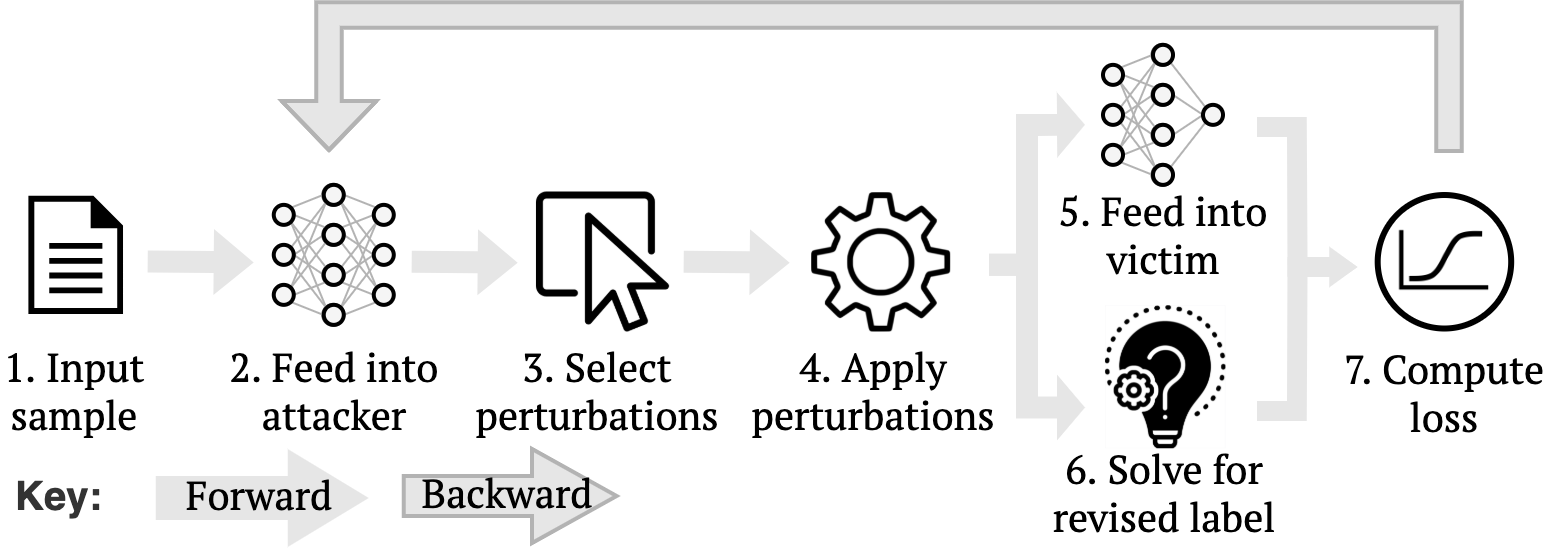}
        \caption{An overview of LAVA. The attacker predicts which perturbations to apply.
        The perturbed sample is fed into the victim and its loss is computed against the revised label, obtained using the \textit{solver}. This acts as the attacker's learning signal.}
        \label{fig:schematic}
    \end{figure}

    We introduce LAVA as a framework for generating logically consistent adversarial attacks. In \S\ref{sec:prob_def} we define our problem statement and introduce the \emph{attacker} and \emph{victim}, and derive our objective and gradients in \S\ref{sec:obj_func}.
    LAVA's overall flow is illustrated in Fig.~\ref{fig:schematic}. The sample is first fed into the attacker which predicts the perturbations to apply\deleteag{ at sentence-level}, relating to stages 2 and 3 from Fig.~\ref{fig:schematic} and \S\ref{sec:architecture}. The perturbed sample is fed into the victim (stage 5). Logical consistency is ensured by recomputing the label for the perturbed sample, relating to step 6 in Fig.~\ref{fig:schematic} and \S\ref{sec:logically_consistent_attacks}, and this is used to compute the attacker's learning signal (stage 7 in Fig.~\ref{fig:schematic}).
    
\subsection{Problem Definition}
    \label{sec:prob_def}

    STP 
    aims 
    to predict if 
    a context, $c$, entails,  $y$, a query, $q$:
    %
    \begin{align}
        \underset{\theta_\mathcal{V}}{\text{max}}\ 
        \log p_{\theta_\mathcal{V}}(y\ |\ q,c) \quad \text{with} \quad y = \mathds{1}(c \models q), \nonumber
    \end{align}
    %
    where 
    $\mathds{1}$ is the indicator function and $p_{\theta_\mathcal{V}}$ is a probability distribution over the entailment label. 
    We refer to the STP, parameterized with $\theta_\mathcal{V}$ (a Transformer), as the \textbf{victim}, as this is the model we are attacking. 
    We probe the victim using adversarial attacks, 
    perturbing the query, $q'$, and/or context, $c'$, \deleteag{in order } \AG{so as} to fool the victim into predicting the opposite label, $\hat{y} = 1 - y = 1 - \mathds{1}(c' \models q')$.
    %
    We train an \textbf{attacker}, parameterized by $\theta_\mathcal{A}$, to generate\deleteag{ these attacks automatically via} victim-fooling perturbations, $(q'_i,c'_i)$, conditioned on the original sample. We represent this as the distribution $p_{\theta_\mathcal{A}}(q_i', c_i'\ |\ q, c)$, \deleteag{dropping subscripts and abbreviating} abbreviated to $p_{\theta_\mathcal{A}}(q', c')$\deleteag{ for brevity}. The objective now is to sample (denoted $\sim$) perturbations from the attacker's distribution to maximize the log likelihood of fooling the victim
    :
    \begin{align}
        \underset{\theta_\mathcal{A}}{\text{max}}\ \log p_{\theta_\mathcal{V}}(\hat{y}\ |\ q',c')\quad \text{where}\quad (q',c') \sim p_{\theta_\mathcal{A}}(q', c').
        \label{eq:adv_objective}
    \end{align}
    Note that we only train the attacker's parameters and the victim is solely used for querying its probabilities (without gradient access), making this a \textit{black-box} adversarial attack method \AG{suitable with any text-based binary classifier}. \deleteag{\textcolor{blue}{LAVA is agnostic to the choice of victim model and \FT{can work with} any text-based binary classifier\delete{ is suitable}.}}
    
    

\subsection{Deriving the Objective Function}
    \label{sec:obj_func}

    \begin{figure*}
        \centering
        \includegraphics[width=1.0\textwidth]{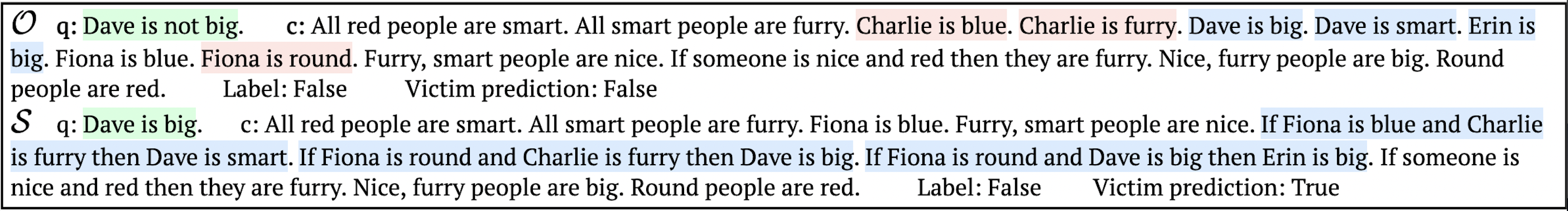}
        \caption{A randomly sampled successful attack, $\mathcal{S}$. The perturbations applied to the original $\mathcal{O}$ are coloured green, red and blue corresponding to QuesFlip, SentElim and EquivSub strategies respectively.}
        \label{fig:adv_example}
    \end{figure*}
    We now introduce the perturbations, $q'$ and $c'$, as latent variables. We seek to maximize the (log) probability of incorrect victim answers, $J = \log p_{\theta_{\mathcal{V}}}(\hat{y}\ |\ q,c)$. Marginalizing over the perturbations, we obtain
    \begin{align}
        J&= \log \sum_{q',c'} p_{\theta_{\mathcal{V}}}(\hat{y}\ |\ q',c')p_{\theta_\mathcal{A}}(q',c'\ |\ q,c) \\
        &= \log \underset{p_{\theta_\mathcal{A}}(q',c')}{\mathbb{E}} \left[ p_{\theta_{\mathcal{V}}}(\hat{y}\ |\ q',c') \right],  \nonumber
    \end{align}
    where perturbations are sampled from the attacker's joint distribution, as outlined in Eq.~\ref{eq:adv_objective}. Following \cite{nvil}, we derive a lower bound using Jensen's inequality, which says that for any convex function $g(x),\ \mathbb{E}[g(x)] \geq g(\mathbb{E}[x])$ \cite{variational_jordan}:
    \begin{align}
        J&\geq \underset{p_{\theta_\mathcal{A}}(q',c')}{\mathbb{E}} \left[\log p_{\theta_\mathcal{V}}(\hat{y}\ |\ q',c') \right].  \nonumber
    \end{align}
    Expanding the expectation gives our \textbf{final optimisation problem}, $\mathcal{L}$:
    \begin{align}
        \underset{\theta_\mathcal{A}}{\text{max}}\ \mathcal{L} &= \sum_{q',c'} p_{\theta_\mathcal{A}}(q', c') \log p_{\theta_\mathcal{V}}(\hat{y}\ |\ q',c') \label{eq:training_obj}\\
        &s.t.\ \hat{y} = 1 -  \mathds{1}(c' \models q'). \nonumber
    \end{align}
    From Eq.~\ref{eq:training_obj} we obtain an expression for the attacker's gradients. We subtract a baseline $b$ to reduce the variance of the learning signal without affecting its expectation:
    \begin{align}
        \nabla_{\theta_\mathcal{A}} \mathcal{L} = \sum_{q', c'} \left( \log p_{\theta_\mathcal{V}}(\hat{y}\ |\ q',c') - b\right) \nabla_{\theta_\mathcal{A}} p_{\theta_\mathcal{A}}(q', c'). \label{eq:grads}
    \end{align}
    Eq.~\ref{eq:grads} relates to the backward pass component of Fig.~\ref{fig:schematic}. As in \cite{nvil}, the baseline is an exponentially-decaying moving average of $\log p_{\theta_\mathcal{V}}(\hat{y}\ |\ q',c')$.
    
\subsection{Attack Strategies}
    \label{sec:attack_strats}
    
    Generating textual adversarial attacks is challenging as there is no universal definition for measuring semantic distances between texts, making it difficult to quantify the distributional shift introduced during attacks. Controlling which perturbations preserve the sample's label is another challenge
    . These are often addressed by assuming locality, namely that several word/sub-word/character perturbations are unlikely to materially impact the text's semantics \cite{sage}.
    This assumption is invalid for logical entailment as this is sensitive even to local perturbations - the \textit{logical inconsistency} problem. As for text generation in general, ensuring that outputs are fluent 
    and coherent 
    is also a consideration
    .
    
    We address these challenges using a structured generative process, specifically by constraining LAVA to select from a set of perturbations. These perturbations were chosen to exploit the logical semantics underlying the data, probing the victim's robustness to logical transformations such as negation. These are implemented using deterministic rules which we refer to as templates. \delete{This} \FT{The use of these perturbations}  ensures the attacks are linguistically \deleteag{consistent}\AG{aligned} with the original data distribution, while also guaranteeing grammaticality
    and fluency. 
    The remainder of this section outlines the perturbations candidates (
    stage 4 in Fig.~\ref{fig:schematic}),  summarised in Table~\ref{tab:attacks}.\deleteag{ Fig.~\ref{fig:adv_example} \delete{contains} \FT{shows} a successful LAVA attack, annotated to show each perturbation type.} These are annotated on a successful LAVA attack in Fig.~\ref{fig:adv_example}.
    
    \begin{table}[b]
    \centering
    \small
    \begin{tabular}{lcc}
        \toprule
        \bf{Perturb.} & \bf{Before} & \bf{After} \\
        \midrule
        QuesFlip: & $q, \{a,b,a \rightarrow c\}$ & $\textcolor{blue}{\lnot} q, \{a,b,a \rightarrow c\}$ \\
        SentElim: & $q, \{a,b,a \rightarrow c\}$ & $q, \{\Ccancel[blue]{a},b,\Ccancel[blue]{a \rightarrow c}\})$ \\
        EquivSub: & $q, \{a,b,a \rightarrow c\}$ & $q, \{\Ccancel[blue]{a},b,a \rightarrow c,\textcolor{blue}{b \rightarrow a}\}$ \\
                \bottomrule
    \end{tabular}
    \caption{Illustration of the perturbations on a logic program with query $q$ and context $\{a,b,a \rightarrow c\}$.}
    \label{tab:attacks}
    \end{table}

    \paragraph{Question flipping (QuesFlip)} Predict whether to negate the question. \cite{negated_probes} motivate this by showing that neural models struggle with \delete{negated reasoning} \FT{negation}.
    To illustrate the use of templates, the sentence \emph{``The \textbf{entity} is \textbf{attribute}''} 
    may be mapped into: \emph{``The \textbf{entity} is not \textbf{attribute}''}.
    
    \paragraph{Sentence elimination (SentElim)} Predict whether to retain each context sentence in the perturbed output. 
    
    \paragraph{Equivalence substitution (EquivSub)} Predict whether each context fact should be substituted with a logically equivalent propositional rule where the fact is the rule head and the body consists of other (randomly sampled) context facts. 
    
\subsection{Attacker Architecture}
    \label{sec:architecture}
    The attack model outputs a categorical distribution with $2N^c + 1$ parameters, $N^c$ for EquivSub and SentElim and one for QuesFlip, where $N^c$ is the number of context sentences. An attack is created by sampling\deleteag{ a set of} perturbations and applying them to the input\deleteag{ sample,} whilst recomputing the associated label. The perturbed output is fed into the victim.
    
    
    The attacker's architecture is similar to \cite{prover}, differing mainly through our three output heads, $g^{1}, g^{2}\ \text{and}\ g^{3}$,  for the attack strategies in \S\ref{sec:attack_strats}. \delete{We form the input
    as $x = [CLS]\ q\ [SEP]\ c\ [SEP]$, with\delete{ tokenized form}\AG{ ids} $\textbf{x}$. This}\AG{The input, $x = [CLS]\ q\ [SEP]\ c\ [SEP]$, with ids $\textbf{x}$,} is fed into a Transformer encoder, $\text{E}_\mathcal{A}$, giving hidden states $\textbf{H} = \text{E}_\mathcal{A}(\textbf{x})$.
    We denote the sequence of tokens and hidden states corresponding to sentence $j$ as $c_j$ and $\textbf{H}_j$ respectively with $c_0 \coloneqq q$. The representation, $\textbf{h}_j$, is formed using mean pooling as
    $\textbf{h}_j = \text{mean\_pool}(\textbf{H}_j)$.
    This is fed into the output heads, $g^{i}$, to obtain the categorical parameter, $\mu^i_j$:
    \begin{align}
        \mu^i_j &= g^i \left( \textbf{h}_j \right) = \sigma \left( \textbf{W}^i_1 \text{ReLU} \left( \textbf{W}^i_2\textbf{h}_j + \textbf{b}^i_2 \right)  + b^i_1 \right). \label{eq:categorical_param}
    \end{align}
    $\textbf{W}_1, \textbf{W}_2, b_1, \textbf{b}_2$ are learnable weights and $\sigma(z) = 1 / (1 + e^{-z})$. The question representation, $\textbf{h}_0$, is only fed into the QuesFlip head, $g^{3}$, while other representations, $\textbf{h}_1, \ldots, \textbf{h}_{N^c}$, are fed into the SentElim and EquivSub heads, $g^{1}$ and $g^{2}$.
    
    The attacker is trained using REINFORCE \cite{REINFORCE}, with the victim's loss as the learning signal. \deleteag{Importantly, the model loss, used to obtain the attacker's gradients, is computed using the logically consistent label}\AG{Crucially, the logically consistent label is used to compute the loss and attacker's gradients}; 
    else the attacker may learn to produce inconsistent attacks. This relates to stage 6 in Fig.~\ref{fig:schematic} and \S\ref{sec:logically_consistent_attacks}.
    
\subsection{Logically Consistent Attacks}
    \label{sec:logically_consistent_attacks}
    
    
    To address the logical consistency challenge outlined in \S\ref{sec:attack_strats}, we must solve for the perturbed sample's entailment relationship ($c' \models q'$ in Eq.~\ref{eq:training_obj}).
    \deleteag{We observe that the entailment label is predictable under the EquivSub and QuesFlip perturbations, with the former preserving the label and the latter flipping it. However}
    While the entailment label is predictable under the EquivSub and QuesFlip perturbations (the former preserves and the latter flips the label), the impact of applying SentElim is unpredictable. Our solution is \delete{to design a }the \textit{solver} module which uses a symbolic solver  to recompute the label for the perturbed sample, denoted the \textit{modified} label. LAVA uses ProbLog \cite{problog} as the solver \FT{but} could be easily adapted to use an alternative \deleteag{solver }if desired. We \delete{are able to }apply the perturbations to the natural language sentences as well as the underlying 
    logic programs which were used to generate them. The perturbed logic program is then fed into the \textit{solver} module to compute the modified label. This occasionally ($< 2\%$) fails due to negative cycles in the perturbed  sample, in which case we denote the attack as unsuccessful.
    
\section{Experiments}
    \label{sec:experiments}
    To our knowledge, LAVA is the first to use logically consistent attacks. 
    \FT{Here, we first introduce}  baselines and metrics 
    \FT{(\S\ref{sec:benchmarks} - \ref{sec:metrics})}. Next, we validate 
    \FT{LAVA}'s effectiveness by showing that 
    \FT{it} successfully generates attacks (\S\ref{sec:attacker_performance}) which transfer across victims 
    and address the victims' weaknesses when trained upon (\S\ref{sec:transf}). All results are 
    for the RuleTakers test set (20,192 samples), using the validation set for early stopping.
    
\subsection{Benchmarks}
    \label{sec:benchmarks}

    \paragraph{HotFlip (HF)} This is a white-box attack method which uses the victims' gradients to estimate the loss of substituting words/characters in the input text, selecting the highest-impact perturbation \cite{hotflip}. For comparability, we use the black-box OpenAttack variant \cite{openattack}, substituting words with synonyms and replacing gradient-guided perturbation search with brute force.
    
    \paragraph{TextFooler (TF)} This is a black-box  method which generates attacks by substituting input words with synonyms
    and ranking them by impact, verifying that the substitutions maintain grammaticality and semantic similarity \cite{textfooler}. We use the OpenAttack \cite{openattack} implementation.

    As mentioned in \S\ref{sec:logically_consistent_attacks}, perturbing the input can compromise a sample's logical consistency, resulting in invalid attacks. We thus integrate the \textit{solver} module into the above 
    HF and TF
    , modifying the logic programs and computing a label for the perturbed sample, as in \S\ref{sec:logically_consistent_attacks}. We report the logical consistency-adjusted attack success rates below.
    
    \paragraph{Random Selector (RS)} This sets $\mu^i_j = 0.5$ from Eq.~\ref{eq:categorical_param}. This ablates LAVA's learned component and makes its perturbations independent of the input, providing a comparison against an attacker that selects perturbations randomly.
    
    \paragraph{Unigram Selector (US)} This is a heuristic version of RS. Perturbation probability is biased towards sentences with greater word overlap with the question, replacing Eq.~\ref{eq:categorical_param} with
    \begin{align}
        \mu^i_j = 0.5 - \bar{r} + r_j,\ \ \text{where}\ \ r_j = \text{ROUGE-1}(c_0, c_j). \nonumber
    \end{align}
    ROUGE-1 \cite{rouge} computes the unigram overlap score between context sentence $c_j$ and the query, $c_0$. $\bar{r}$ is the mean unigram score over the training data, giving the same expected categorical parameters as under RS.
    
\subsection{Metrics}
    \label{sec:metrics}

    \begin{table}
    \centering
    \small
    \begin{tabular}{llrr}
    \toprule
    & \bf{Name} & \bf{ASR (\%)} $\uparrow$ & \bf{F1 (\%)} \\
    \midrule
    \parbox[t]{2mm}{\multirow{4}{*}{\rotatebox[origin=c]{90}{\bf Baselines}}} & TF & 10.1 &  88.9 \\
    & RS  &    10.4 &  66.9 \\
    & US  &    12.1 &  65.3 \\
    & HF & 21.5 &  94.5 \\
    \midrule
    \parbox[t]{2mm}{\multirow{3}{*}{\rotatebox[origin=c]{90}{\bf Abls.}}}
    & $\mathcal{R}_l -$ SentElim & 10.0 &  74.3 \\
    & $\mathcal{R}_l -$ EquivSub & 10.9 &  80.3 \\
    & $\mathcal{R}_l -$ QuesFlip & 25.1 &  68.4 \\
    \midrule
    \parbox[t]{2mm}{\multirow{3}{*}{\rotatebox[origin=c]{90}{\bf Ours}}} & \bf $\mathcal{R}_l$ (main) &    28.6 &  67.6 \\
    & $\mathcal{R}_b$ &    30.1 &  67.8 \\
    & $\mathcal{R}_d$ &    52.5 &   68.6 \\
    \bottomrule
    \end{tabular}    
    \caption{Attacker results. Our main model (used throughout this study) uses \textsc{RoBERTa}-large, $\mathcal{R}_l$, as the victim. We provide \textsc{RoBERTa}-base, $\mathcal{R}_b$, and Distil-\textsc{RoBERTa}-base, $\mathcal{R}_d$, as references as these are weaker \FT{(easier to attack)} victims
    , hence ASR is higher. The baselines 
    \FT{(see} \S\ref{sec:benchmarks}\FT{)} and ablations \FT{(Abls.)} use $\mathcal{R}_l$ as the victim so should be compared to $\mathcal{R}_l$ \textbf{(main)}.
    The ablations 
    begin from $\mathcal{R}_l$ \textbf{(main)} and show which attack strategy was removed.
    Higher is better for ASR, but not necessarily for F1, as per \S\ref{sec:metrics}.
    }
    \label{tab:attacker}
    \end{table}

    \paragraph{Attack success rate (ASR)}  This represents the attacker's performance: ASR = \# successful attacks / \# attacks.

    \paragraph{F1 sentence overlap score (F1)} This is a measure of the distance between perturbed and original examples in terms of sentence overlap. Higher scores reflect more overlap and while a low F1 is bad (``adversarial''\delete{ attacks should use } \AG{implies} minimal perturbations), we do not seek to maximise F1 as this can overly constrain the attacker and restrict \delete{the diversity of its attacks. In other words,} \FT{attacks' diversity (thus,} a higher value for F1 is not necessarily preferable\FT{)}.

    Prior studies \cite{seqattack,bert-attack} report grammaticality and fluency, but these are unnecessary here due to our template-based generative process.
    
\subsection{Attacker Performance}
    \label{sec:attacker_performance}



    Before training our attacker, we train the victims until convergence on the RuleTakers dataset. Unless specified, the victim's core uses \textsc{RoBERTa}-large, $\mathcal{R}_l$, \cite{roberta} and we provide \textsc{RoBERTa}-base, $\mathcal{R}_b$, and Distil-\textsc{RoBERTa}-base, $\mathcal{R}_d$, \cite{distilroberta} for comparison. The attacker is trained for five epochs with a batch size of 8 on a single 11Gb NVIDIA GeForce RTX 2080 GPU. The learning rate was set to 5e$^{-6}$, with 1e$^{-5}$ and 2.5e$^{-6}$ also tested. 
    We provide configuration files in the accompanying code for reproducibility.

    \paragraph{ASR} Table~\ref{tab:attacker} gives the attackers' results. We report the four benchmarks from \S\ref{sec:benchmarks} and model ablations. For HF and TF, we report the adjusted results, described in \S\ref{sec:benchmarks}, to ensure logical consistency. The unadjusted ASRs are 81\% and 84\% (not shown in the table) respectively, vs 22\% and 10\% after adjusting. \AG{Recall that the unadjusted figures contain many logically-inconsistent attacks, hence we only discuss the adjusted ASR.} This demonstrates that naively perturbing the context will likely flip the entailment relationship, hence integrating the \textit{solver} \deleteag{module into }\AG{within} the generative process is desirable. This result corroborates recent findings that adversarial methods\delete{ are less effective after removing} \AG{often generate} invalid examples \cite{human_vs_machine_adv}. Using an independent t-test, our method significantly outperforms all\delete{ (adjusted)} baselines. The F1 metric is discussed in \S\ref{sec:num_preturbs}.
    
    
    
    \paragraph{Best-of-$k$ Decoding} The results in Table~\ref{tab:attacker} use \textit{one-}
    \begin{wrapfigure}{r}{4.5cm}
        \centering
        \vspace{-0.3cm}
        \includegraphics[width=4.1cm]{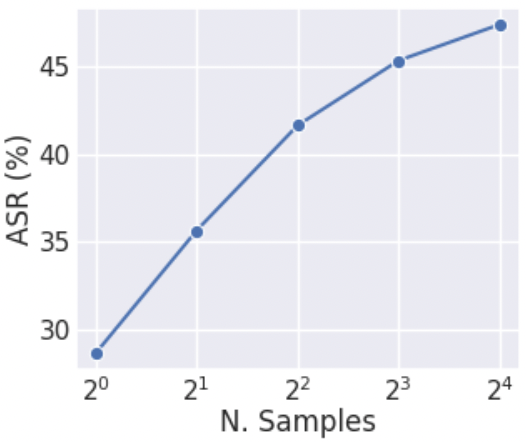}
        \vspace{-0.3cm}
        \caption{ASR vs $\mathcal{N}^k$.}
        \vspace{-0.3cm}
        \label{fig:val_mc_vs_asr}
    \end{wrapfigure}
    \textit{sample} decoding: the attacker generates one attack for each sample. Alternatively, we can use \textit{best-of-$k$ decoding} by taking $\mathcal{N}^k$ adversarial samples and retaining the one which most successfully fools the victim. Fig.~\ref{fig:val_mc_vs_asr} shows the benefits of best-of-$k$ decoding, yielding an ASR of 47\% when $\mathcal{N}^k = 16$.

    The $\mathcal{R}_l$ victim performs near-perfectly on the original task with 99.3\% accuracy (reported in Table~\ref{tab:adv_retraining}, discussed in \S\ref{sec:adv_training}). Given this, it is interesting that the attacker fools the victim with 29\% and 49\% ASR when $\mathcal{N}^k = 1$ and $16$ respectively. This reveals weaknesses in the victim's reasoning capabilities which the attacker learns to exploit.

    \paragraph{Ablations} Here we ablate each perturbation strategy from \S\ref{sec:attack_strats}
    and compare this against the performance when using all three ($\mathcal{R}_l$ \textbf{(main)} in Table~\ref{tab:attacker}). 
    Using a t-test, all ablations significantly reduce the attacker's ASR ($p < 3e^{-15}$). Equipping the attacker with a greater range of perturbations helps it to identify weaknesses in the victim's reasoning capabilities. 

\subsection{Transferability}
    \label{sec:transf}

    \begin{table}
    \centering
    \small
    \begin{tabular}{clrr}
    \toprule
    \bf Method & \bf Victim & \bf Trf. ASR (\%) $\uparrow$ & \bf Atk. ASR (\%)\\
    \midrule
    \multirow{4}{*}{\bf Ours} & $\mathcal{R}_l \rightarrow \mathcal{R}_b$ & 84.4 & 28.6 \\
    & $\mathcal{R}_l \rightarrow \mathcal{R}_d$ & 82.1 &         28.6 \\
    & $\mathcal{R}_b \rightarrow \mathcal{R}_d$ & 81.8 &         30.1 \\
    & $\mathcal{R}_b \rightarrow \mathcal{R}_l$ & 80.2 & 30.1 \\
    \midrule
    \textbf{TF} & $\mathcal{R}_l \rightarrow \mathcal{R}_b$ & 64.6 & 10.1 \\
    \textbf{HF} & $\mathcal{R}_l \rightarrow \mathcal{R}_b$     & 55.1 & 21.5 \\
    \bottomrule
    \end{tabular}
    \caption{The transferability, \textit{Trf.}, of adversarial samples.  $\mathcal{V}_x \rightarrow \mathcal{V}_y$ means the attacks were generated against victim $\mathcal{V}_x$ and are evaluated against $\mathcal{V}_y$. Attacker, \textit{Atk.}, ASR is shown for comparison.
    }
    \label{tab:transferability}
    \end{table}
    
    In general, when a model is used to generate attacks, a concern is that the attacker can overfit to a single victim. 
    This can be formalized using \textit{transferability} \cite{ZOO}, defined as the proportion of the successful attacks generated against victim $x$, denoted $\mathcal{V}_x$, that also successfully fool a different victim $y$, denoted $\mathcal{V}_y$,
    shown in Table~\ref{tab:transferability}. \delete{The top row from Table~\ref{tab:transferability} says that 84\% of the attacks that were successful when $\mathcal{R}_l$ is the victim also succeed when $\mathcal{R}_b$ is the victim.}In Table~\ref{tab:transferability}, the top row says that 84\% of the successful attacks against $\mathcal{R}_l$ also succeed against $\mathcal{R}_b$. Our methods' attacks are more transferable than the baselines, implying that LAVA identifies global rather than local vulnerabilities.

\subsection{ASR vs Number of Perturbations}
    \label{sec:num_preturbs}
    
    \begin{figure}[b]
        \centering
        \includegraphics[width=7cm]{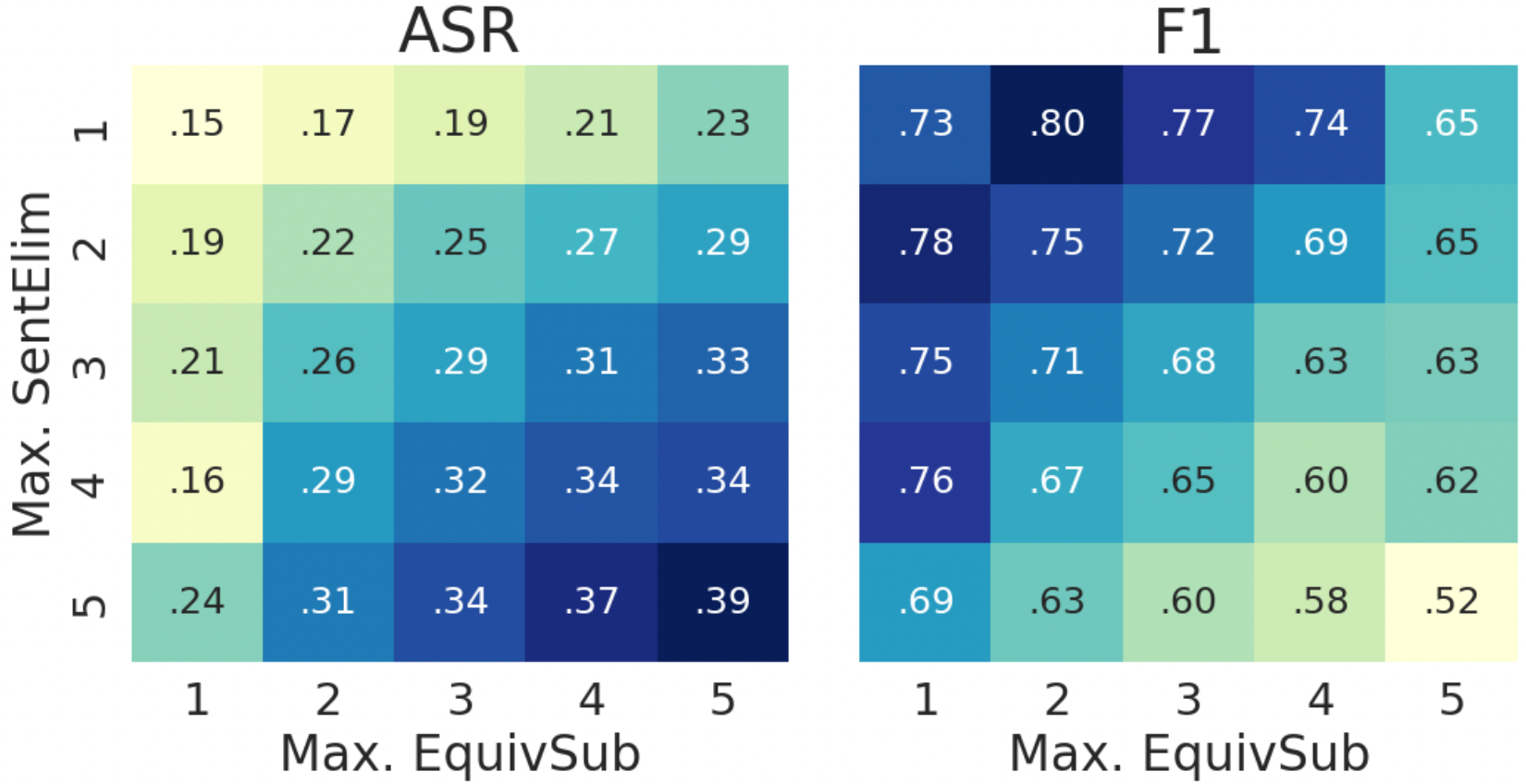}
        \vspace{-0.2cm}
        \caption{Heatmaps depicting the impact of the maximum SentElim and EquivSub perturbations on ASR and F1 score.}
        \label{fig:num_perturbs}
    \end{figure}
    
    The results according to the two metrics, ASR and F1, cannot be taken in isolation, since they encapsulate an important trade-off between the ease of fooling the victim and the number of applied perturbations.
    In our early experiments we observed that the attacker could easily achieve a high ASR by 
    perturbing most of the context sentences. This shifts the original and perturbed samples' distributions, making it trivial to fool the victim but hard to attribute its failure to poor reasoning versus out-of-distribution generalization.
    Addressing this, we introduce hyper-parameters capping the number of SentElim and EquivSub perturbations the attacker can apply. Fig.~\ref{fig:num_perturbs} shows the trade-off between ASR and F1: increasing this threshold leads to more successful attacks but also makes the original and perturbed samples less similar. Note that these parameters specify the maximum number of perturbations; the attacker can apply fewer. Interestingly, our main model (with the caps set at 3) uses 2.3 EquivSub perturbations on average. As EquivSub adds a reasoning step, thus in principle making the sample more difficult, we expected the attacker to use EquivSub as frequently as permitted.
    
    The results reported elsewhere in this study assume the SentElim and EqivSub caps are set at 3, the level we consider to optimally trade-off ASR and F1. In Table~\ref{tab:attacker} the F1 scores of HF and TF are near 1, reflecting the local nature of these methods' attacks which limits their diversity. This metric does not capture the fact that logically equivalent sentences can be realised linguistically distinctly, thus requiring a high F1 score can overly constrain the attacker. 

\subsection{Adversarial Training}
    \label{sec:adv_training}
    The previous sections demonstrated that an attacker can learn to exploit weaknesses in the victim's reasoning capabilities\deleteag{ via a few, basic perturbations}. Here, we investigate whether training the victim on the successful attacks helps address these blind spots. To test this, we initially train a victim to convergence on the original dataset $\mathcal{D}^O$ and train an attacker against this victim. The attacker is evaluated on $\mathcal{D}^O$, generating an augmented dataset of adversarial samples, $\mathcal{D}^A$. We then resume training the victim on the combined dataset $\mathcal{D}^C = \mathcal{D}^O \bigcup \mathcal{D}^A$.
    
    The results in Table~\ref{tab:adv_retraining} show that training on $\mathcal{D}^C$ improves the victims' performance on $\mathcal{D}^O$ and $\mathcal{D}^A$. Using an independent t-test, all differences are significant with $p < 1e^{-7}$. While there is limited headroom for improvement on $\mathcal{D}^O$ using $\mathcal{R}_l$ and $\mathcal{R}_b$ as victims, the weaker distilled model, $\mathcal{R}_d$, sees a sizable 17.5\% improvement. These show that data augmentation via logically consistent adversarial examples is a promising strategy for enhancing STPs. 

    \begin{table}
    \centering
    \small
    \begin{tabular}{ccrrr}
    \toprule
    \bf Victim & \bf DSet & \bf Before & \bf After $\uparrow$ & \bf $\Delta$ (\%) $\uparrow$ \\
    \midrule
    \multirow{2}{*}{$\mathcal{R}_l$} & $\mathcal{D}^A$ &   71.4 &  96.9 & 35.7 \\
          & $\mathcal{D}^O$ &   99.3 &  99.7 & 0.4 \\
    \multirow{2}{*}{$\mathcal{R}_b$} & $\mathcal{D}^A$ &   69.9 &  94.9 & 35.8 \\
          & $\mathcal{D}^O$ &   97.9 &  98.6 & 0.7 \\
    \multirow{2}{*}{$\mathcal{R}_d$} & $\mathcal{D}^A$ &   47.5 &  85.4 & 79.7 \\
          & $\mathcal{D}^O$ &   75.7 &  89.1 & 17.7 \\
    \bottomrule
    \end{tabular}
    \caption{Adversarial training results showing the victim's accuracy on the original, $\mathcal{D}^O$, and adversarial, $\mathcal{D}^A$, datasets before and after training on the augmented dataset, $\mathcal{D}^C$.
    }
    \label{tab:adv_retraining}
    \end{table}

\section{Analysis of Attacks}
    \label{sec:analysis}


    

    
    \paragraph{Qualitative} We have identified several blind spots in the victim's reasoning capabilities by qualitatively analysing the successful attacks. Fig.~\ref{fig:adv_example} displays an example of one such vulnerability, whereby the victim is often fooled if the query literal appears within the body of a rule, i.e. ``If Fiona is round and \textit{Dave is big} then ...''. This can be exploited by the attacker's EquivSub perturbation. Another common failure type is when the attacker eliminates a leaf fact on samples containing  
    deeper proofs. This operation flips the label while the victim's prediction is unchanged. This suggests the victim performs backward-chaining from the query, but its search is shallow and terminates before the proof tree is complete.
    
    More generally, the victim appears brittle when using quantified rules. It struggles to correctly bind variables on either side of an implication, e.g. ``if someone is happy then they like John'' versus ``if someone is happy then Anne likes John''. Similarly, many failures occur on conjunctive implications containing both variables and facts. Using the example ``if Anne likes John and someone is happy then they like Beth'', \textit{someone} and \textit{they} refer to a common variable but could be misconstrued as relating to \textit{Anne} or \textit{John}. However, none of the attacker's perturbations can create quantified rules so it cannot exploit these vulnerabilities.
    
    These observations yield intriguing insights into STPs' reasoning processes and suggest avenues for improving the STPs and attackers.
    The first and second failure modes imply that short-cuts are used, although we believe they play a minor role else the victim would prove more brittle. The failures involving quantified rules suggest a coarse word-sense disambiguation mechanism which is not scaling to more complex rules. It also suggests an incomplete understanding of the properties of variables within logic programs.
    
    
    \paragraph{Quantitative} Table~\ref{tab:asr_stratified} shows the attacker's ASR vs proof
    \begin{wraptable}{r}{2.5cm}
        \vspace{-0.2cm}
        \centering
        \small
        \begin{tabular}{lr}
        \toprule
        \bf{Len} & \bf{ASR} $\uparrow$\\
        \midrule
        0              &      0.0 \\
        1              &     19.3 \\
        2              &     30.0 \\
        $\geq$3              &     47.4 \\
        \bottomrule
        \end{tabular}
        \caption{Attacker performance (ASR, \%) by proof length.
        }
        \label{tab:asr_stratified}
    \end{wraptable}
    length. Predictably, as samples get harder the victim is more easily fooled. Notice that for proofs of length zero (20\% of samples) the attacker's ASR is zero. 
    By definition, proofs of length zero have no context sentences \delete{which unify}\FT{unifying} with the query (it does not ``match'' any facts or rule heads), rendering attacks ineffective as the attacker cannot synthesise a fact\delete{ or }\FT{/}rule to link the query and the context.
    
    
    
    

\section{Related Work}
    \label{sec:related_work}

    \paragraph{Transformers as Soft Theorem Provers} Our study is closely related to the \textit{RuleTakers} line of work \cite{ruletaker}. These posit that Transformers can act as ``soft theorem provers'' over natural language sentences, analog
    \FT{ous} to symbolic theorem provers for formally represented theories.
    These studies exhibit impressive model capabilities, such as near-perfect entailment prediction, an ability to generate reasoning proofs \cite{prover}, generalizing beyond the training set 
    and reasoning over implicit and explicit knowledge \cite{leapofthought}. Our work stands apart by identifying weaknesses in these approaches using adversarial attacks.
    
    Other studies have noted the shortcomings of Transformers on reasoning problems. Vulnerability to deeper Horn Rule reasoning \cite{systematicity}, word-order permutations \cite{right_for_wrong_reasons}, mis-priming \cite{mispriming} and shallow heuristics \cite{shallow_heuristics} have all been observed (see \cite{helwe2021reasoning} for more examples). Our study is distinct from these as we train a generative adversary end-to-end to discover these weaknesses.

    \paragraph{Adversarial Attacks} 
    These have seen recent attention by the NLP community as a method 
    for probing model robustness and creating more challenging datasets. Borrowing the taxonomy\delete{from} \FT{in} \cite{seqattack}, these can \FT{be} grouped according to: 1) specificity - are the attacks targeted at a label\FT{?} 2) Knowledge of the victim - \textit{white-box} attacks assume all information about the victim is known,
    whereas \textit{black-box} attacks assume only its probabilities are known. 3) Granularity - are characters, tokens or phrases perturbed? We add a fourth \textit{data generation process} criterion \delete{to these} - is it solely \FT{a} human, a hybrid human-in-the-loop or a solely model-based procedure? Using this taxonomy, our model is an unspecified,  black-box, phrase-level, model-based attack procedure. 
    
    Several \delete{popular} methods have recently emerged at \delete{the} character-level \cite{hotflip,deepwordbug} and word-level \cite{bert-attack,textfooler}. These generally rely on inserting/removing characters or substituting words with synonyms or tokens with similar embeddings. These methods implicitly assume that the locality of attacks preserves the semantics of the original passage. \delete{More recently,} This assumption has \FT{recently} proven invalid \cite{human_vs_machine_adv}, particularly for logical reasoning we posit as entailment is highly sensitive to token-level perturbations.
    
    \delete{Our work} 
    \FT{LAVA} was inspired by a recent movement towards training generative models for adversarial attacks. Ours is closely related to SAGE \cite{sage}, a Wasserstein Auto-Encoder trained to attack TableQA systems. A particular challenge in this domain is controlling the semantic shift introduced by attacks \cite{human_vs_machine_adv} - logical consistency can be viewed as a special case of semantic shift. \delete{SAGE's focus on semantically-valid attacks inspired our emphasis on logical consistency.} Other generative strategies have been successfully employed, such as using GANs \cite{NL_adversarial} or integrating victim gradients \cite{GBDA}. 


\section{Conclusions and Future Work}
    We use adversarial examples to probe the robustness of STPs. To this end, we propose LAVA, a black-box attack framework which learns to exploit vulnerabilities in the victim's reasoning capabilities. We define the \textit{logical inconsistency} problem, namely that a given sample's logical entailment relationship may be sensitive to even small perturbations, and we address this via a structured generative process in conjunction with a integrated symbolic solver. LAVA outperforms standard methods in terms of attack success rate and transferability and training the victim on the adversarial samples improves performance. Our analyses reveal weaknesses and crude heuristics in the victim's reasoning processes, exposing a flawed grasp of the semantics of  logic programs. Future work may seek to equip the attacker with the ability to synthesise rules as we expect this would expose a wider range of vulnerabilities. Going forward, we envisage that LAVA could play a useful role in future STP development as a framework for identifying and debugging their vulnerabilities.

    
\section*{Acknowledgments}
    This work was supported by UK Research and Innovation [grant number EP/S0233356/1], in the UKRI Centre for Doctoral Training in Safe and Trusted Artificial Intelligence. We thank Alex Spies for his insightful comments and feedback.



\bibliographystyle{named}
\bibliography{main}

\end{document}